\documentclass[10pt,twocolumn,letterpaper]{article}

\usepackage{iccv}              

\usepackage{adjustbox}
\usepackage{color}
\usepackage[dvipsnames]{xcolor}

\usepackage{color}

\definecolor{iccvblue}{rgb}{0.21,0.49,0.74}

\usepackage[pagebackref,breaklinks,colorlinks,allcolors=iccvblue]{hyperref}

\def\paperID{6} 
\def\confName{ICCV}
\def\confYear{2025}

\title{Improving Viewpoint Consistency in 3D Generation via Structure Feature and CLIP Guidance}

\author{Qing Zhang$^{1}$,  Jinguang Tong$^{1,2}$, Jing Zhang$^{1}$, Jie Hong$^{3}$, Xuesong Li$^{1, 2}$\thanks{Corresponding author} \\
\small $^{1}$ The Australian National University \quad $^{2}$CSIRO, \quad $^{3}$ The University of Hong Kong\\
{\tt\small xuesong.li@anu.edu.au}
}

\begin{document}
\maketitle
\begin{abstract}
Despite recent advances in text-to-3D generation techniques, current methods often suffer from geometric inconsistencies, commonly referred to as the Janus Problem. This paper identifies the root cause of the Janus Problem: viewpoint generation bias in diffusion models, which creates a significant gap between the actual generated viewpoint and the expected one required for optimizing the 3D model. To address this issue, we propose a tuning-free approach called the Attention and CLIP Guidance (ACG) mechanism. ACG enhances desired viewpoints by adaptively controlling cross-attention maps, employs CLIP-based view-text similarities to filter out erroneous viewpoints, and uses a coarse-to-fine optimization strategy with staged prompts to progressively refine 3D generation. Extensive experiments demonstrate that our method significantly reduces the Janus Problem without compromising generation speed, establishing ACG as an efficient, plug-and-play component for existing text-to-3D frameworks.

\end{abstract}

\section{Introduction}
\label{sec:intro}
In recent years, industries such as gaming, film production, and architecture have experienced a surge in demand for high-quality 3D assets. However, designing these assets remains a labor-intensive process that requires meticulous planning, detailed modeling, and significant computational resources. While generative models have excelled in text-to-image tasks thanks to annotated image datasets scaling to five billion images~\cite{schuhmann2022laion}, the largest publicly accessible 3D databases contain only tens of millions of entries~\cite{deitke2024objaverse}. This discrepancy naturally raises the question of how to harness 2D data sources to develop robust 3D generation capabilities.
\begin{figure}[t]
    \centering
    \includegraphics[width=\linewidth]{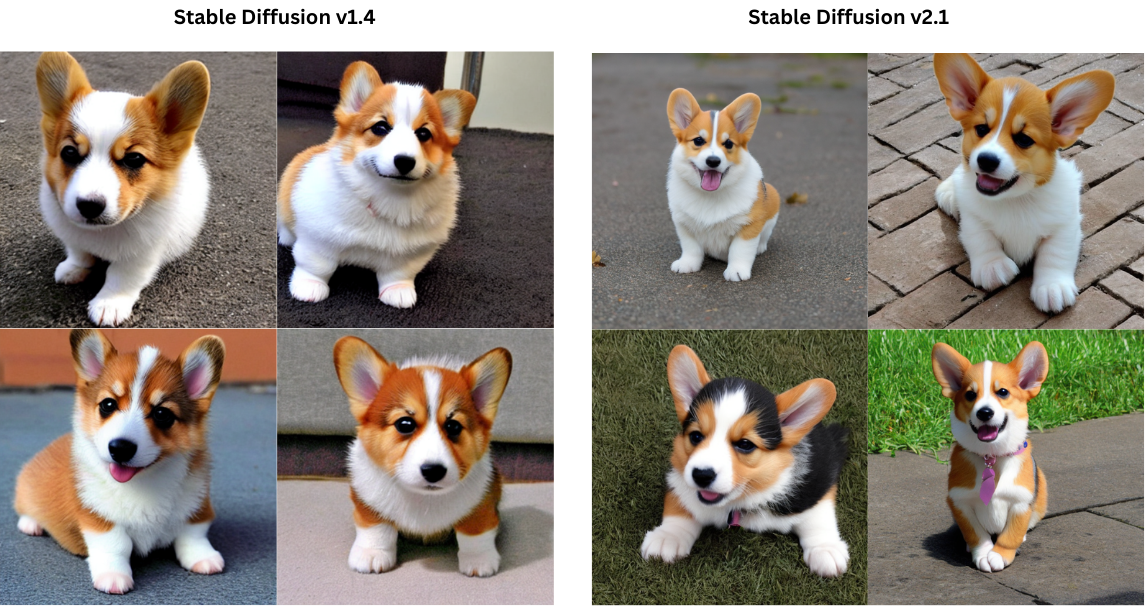}
    \caption{\textbf{Viewpoint bias in Stable Diffusion models.} We show generation samples from Stable Diffusion v1.4 and v2.1 from the prompt “a corgi puppy”. Most samples show a corgi in a forward-facing pose.}
    \vspace{-1.0em}   
    \label{fig:generation_bais}
\end{figure}

DreamFusion~\cite{poole2022dreamfusion} has emerged as a pioneering approach in the realm of 2D lifting to 3D generation. It established a foundational framework by leveraging pre-trained diffusion models to optimize Neural Radiance Fields (NeRF)~\cite{mildenhall2021nerf}, enabling the generation of detailed 3D representations from textual descriptions. Central to this framework is the Score Distillation Sampling (SDS) algorithm, which has demonstrated robust performance across various tasks~\cite{cao2024dreamavatar, chen2023fantasia3d, lin2023magic3d, metzer2023latent, liang2024luciddreamer, poole2022dreamfusion, wang2024prolificdreamer}. Despite its successes, many text-to-3D models that utilize SDS encounter a common issue known as the Janus Problem. As illustrated in Figure~\ref{fig:example}, this problem manifests as multiple views of an object being consistently rendered as front-facing images, leading to geometric inconsistencies and unrealistic 3D models.
\begin{figure*}[t]
    \centering
    \includegraphics[width=\linewidth]{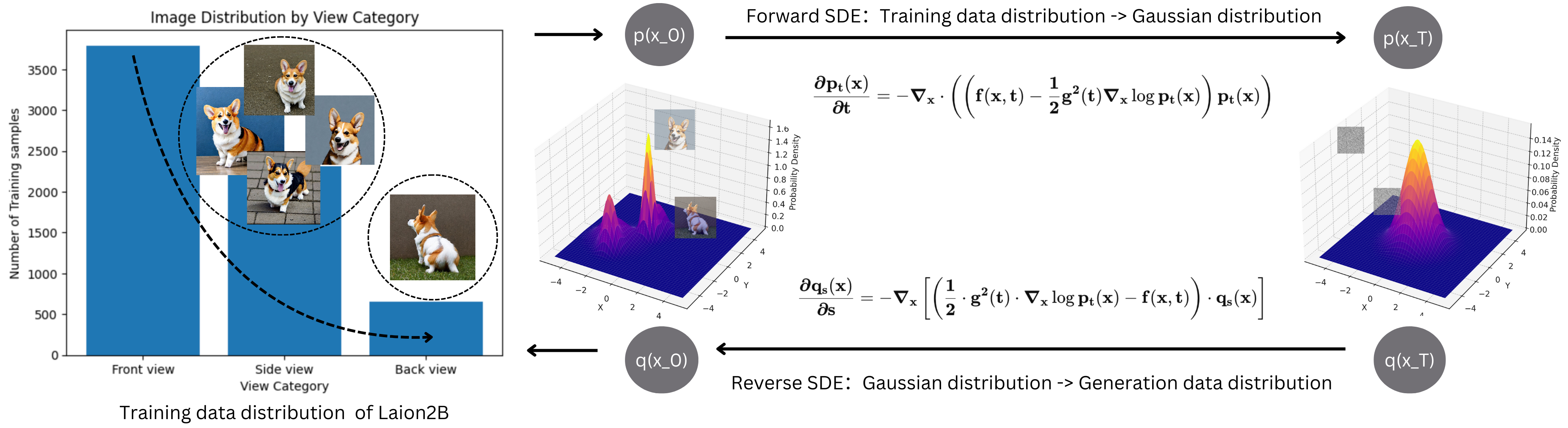}
    \vspace{-0.6em}
    \caption{\textbf{The long-tailed distribution of viewpoints in the diffusion model's training data biases the generated viewpoints.} The left figure shows a random sample of 6,762 images from Laion2B~\cite{schuhmann2022laion}, categorized into three object viewpoints using CLIP ViT-H/14~\cite{ilharco_gabriel_2021_5143773}. The diffusion model maps this long-tailed distribution to a Gaussian distribution, which then produces a similarly biased generation distribution via reverse SDE—fundamentally causing incorrect guidance in 3D representations in the Janus Problem.}
    \vspace{-0.8em}
    \label{fig:long_tail}
\end{figure*}

Previous studies have attributed the Janus Problem to a generative bias inherent in diffusion models~\cite{armandpour2023re, huang2024dreamcontrol, hong2023debiasing, liu2023zero}. Given an input prompt, diffusion models tend to generate front-facing images of objects, contributing to geometric inconsistencies in the resulting 3D models. Some methods attempt to mitigate this issue by fine-tuning diffusion models on specialized 3D datasets to enhance multi-view generation capabilities~\cite{liu2023zero, shi2023mvdream, han2024vfusion3d}. However, directly fine-tuning on a 3D dataset can lead to catastrophic forgetting and overfitting~\cite{hu2021lora, ruiz2023dreambooth, zhang2023adding}, in addition to being time-consuming and computationally intensive. Alternatively, other approaches address the issue by adjusting sampling strategies to achieve more balanced samples~\cite{huang2024dreamcontrol}. Nevertheless, without precise knowledge of the long-tailed distribution of the data, such strategies rely on diffusion models to estimate the distribution during sampling, which is often inaccurate~\cite{zhang2023deep}.

This work demonstrates that the Janus Problem arises from a viewpoint generation bias inherent in diffusion models, which introduces imbalanced guidance for 3D optimization. We identify two primary factors contributing to this bias. First, the long-tailed distribution of training data in diffusion models introduces a bias toward generating images from high-density regions, specifically favoring front views of objects. Second, as input prompt complexity increases, the model’s attention to view-related text diminishes, leading to geometric inconsistencies in the generated 3D models. Together, these factors impede the model’s ability to produce accurate and consistent multi-view representations, thereby exacerbating the Janus Problem.

To address the aforementioned limitations, we propose Attention and CLIP Guidance (ACG), a tuning-free approach designed to resolve the Janus Problem. Specifically, we enhance the model’s focus on viewpoint-related terms by adaptively controlling the cross-attention map within the diffusion model, facilitating the generation of images in low-probability density regions. Furthermore, we leverage CLIP to assess the viewpoints of the model-generated pseudo-Ground-Truth (pseudo-GT), pruning incorrect guidance when mismatched viewpoints are detected, with the aim of rebalancing the distribution of guidance viewpoints introduced to the 3D generation process. Finally, we employ a coarse-to-fine optimization strategy with staged prompts, initially focusing on object geometry and subsequently refining appearance and detailed features. In summary, our contributions are as follows:

\begin{itemize}
    \item We conduct a comprehensive analysis of the Janus Problem in text-to-3D generation, identifying its roots in the long-tailed distribution of the training data and the imbalance in diffusion guidance.
    
    \item We introduce a tuning-free strategy, ACG, to mitigate the Janus Problem through attention and CLIP guidance. ACG functions as a plug-and-play component, making it compatible with and beneficial for existing text-to-3D methods.
    
    \item We conduct extensive comparative experiments against previous methods and perform ablation studies to evaluate the contribution of each module in ACG.
\end{itemize}
\section{Related work}
\label{sec:related_work}
\begin{figure*}[t]
    \centering
    \includegraphics[width=1\linewidth]{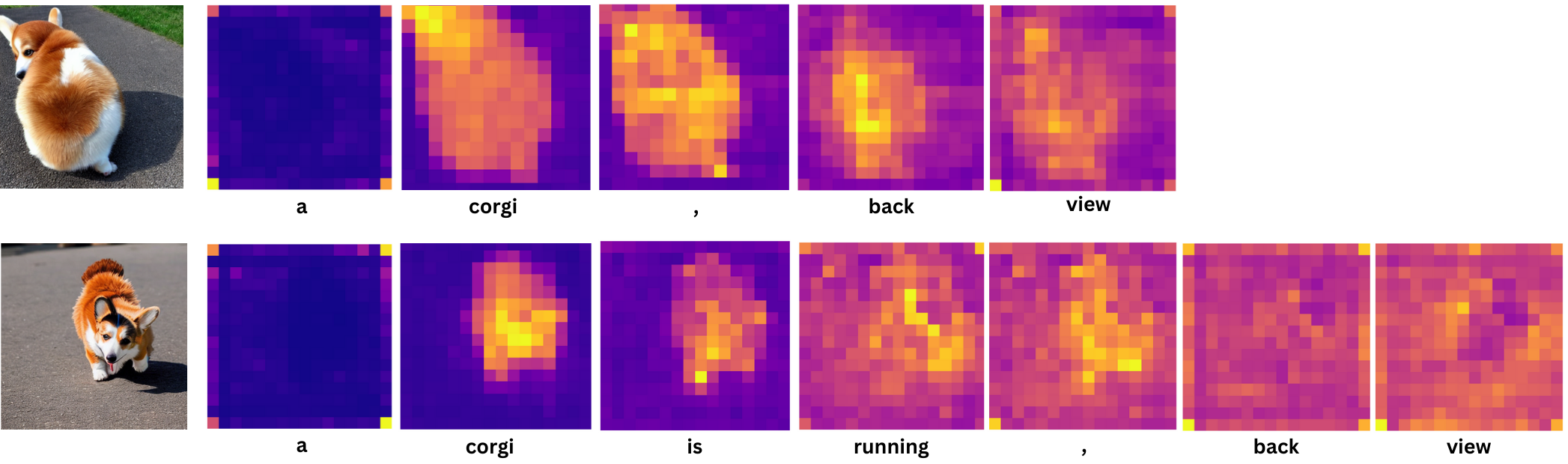}
    \vspace{-1.2em}
    \caption{\textbf{Average attention maps.} As the prompt complexity increases, the diffusion model's attention to viewpoint tokens "back" decreases significantly.}
    \vspace{-1.0em}
    \label{fig:attention_map}
\end{figure*}

\subsection{Text-to-3D Generation}
The task of generating 3D content from text can be broadly divided into two categories. The first category includes end-to-end models such as Point-E~\cite{nichol2022point} and Shap-E~\cite{jun2023shap}, which generate 3D outputs directly from text. These models offer fast generation but suffer from poor generalization and subpar quality due to the limited availability of 3D datasets~\cite{liu2024comprehensive}. The second category follows a 2D-to-3D approach inspired by DreamFields~\cite{jain2022zero}, where CLIP~\cite{radford2021learning} supervises NeRF~\cite{mildenhall2021nerf, xiao2025neural}. However, the limited supervisory effectiveness of CLIP has prevented this approach from gaining widespread adoption. DreamFusion~\cite{poole2022dreamfusion} introduced Score Distillation Sampling (SDS), leveraging pre-trained diffusion models to guide NeRF optimization, which inspired numerous follow-up studies.

Subsequent improvements to DreamFusion have focused on three areas: generation quality, speed, and the Janus problem~\cite{wang2024prolificdreamer,lin2023magic3d, liang2024luciddreamer, chen2023fantasia3d, huang2024dreamcontrol, yi2024gaussiandreamer, han2024latent, chen2024text,tang2023dreamgaussian}. Efforts to improve quality mainly revolve around optimizing SDS, as seen in works like ProlificDreamer~\cite{wang2024prolificdreamer} and LucidDreamer~\cite{liang2024luciddreamer}, which analyze how SDS leads to mode collapse and excessive smoothing. Speed improvements typically involve optimizing the 3D representation and the generation framework. For example, Magic3D~\cite{lin2023magic3d} introduces a two-stage optimization strategy and incorporates DMTet~\cite{shen2021deep} for fast surface optimization, and DreamGaussian~\cite{tang2023dreamgaussian} employs explicit representations to accelerate generation by exploiting the fast rendering speed of Gaussian Splatting~\cite{kerbl20233d, tong2025gs, li2024dgns}. The Janus problem, which concerns generating consistent front and back views, is the primary focus of our research. Previous solutions rely on fine-tuning diffusion models to obtain multi-view capabilities~\cite{liu2023zero, shi2023mvdream}. In contrast, our work takes a different approach by analyzing the inherent generative bias of diffusion models. We achieve superior results without compromising the model's generalization ability.

\subsection{Diffusion Model}
One key component in text-to-3D generation is the diffusion model~\cite{rombach2022high, ho2020denoising}, which provides supervision for the 3D model. The core of generative models is to approximate the distribution of real data, enabling the generation of data consistent with the training set~\cite{kingma2013auto}. Diffusion models focus on mapping training data distribution to a simpler distribution~\cite{ho2020denoising}. Once this mapping is learned, new data can be generated by sampling from the simple distribution and applying the learned mapping. For individual data points, the forward process of diffusion models gradually maps the data to a Gaussian distribution, while the reverse process generates data from the Gaussian distribution:
\begin{equation}
p(x_t|x_0) = \mathcal{N}(x_t; \sqrt{\bar{\alpha}_t} x_0, (1 - \bar{\alpha}_t)I),  \label{equation1}
\end{equation}

\begin{equation}
q_\theta(x_{t-1}|x_t) = \mathcal{N}(x_{t-1}; \mu_\theta(x_t, t), \sigma_t^2 I),  \label{equation2}
\end{equation}
where the term \(\mu_\theta(\cdot)\) represents a function that combines \(x_t\) and a model that predicts the noise \(\epsilon_\theta\) in the data at time step \(t\)~\cite{ho2020denoising}.

Both the forward and reverse processes can be expressed by stochastic differential equations~\cite{song2020score, song2019generative}:

\begin{equation}
dx = f(x,t) \, dt + g(t) \, dw,  \label{equation3}
\end{equation}

\begin{equation}
dx = \left[f(x,t) - g(t)^2 \nabla_{x} \log p_t(x) \right] dt + g(t) \, d\bar{w},  \label{equation4}
\end{equation}
where \( f(x,t) \) is the drift coefficient, \( g(t) \) is the diffusion coefficient, and \( dw \) represents the differential of the Wiener process. In the reverse process, \(\bar{\mathbf{w}}\) is the Wiener process when time flows backward from \(T\) to \(0\)~\cite{song2020score}.

The intuition that generative models approximate the training data led us to question whether the long-tailed distribution of the training data affects the generative capability of diffusion models, thereby providing unreliable supervision to NeRF. Based on this intuition, we validate its rationale in Section~\ref{sec:rethin_Janus}, which led to our method.
\section{Method}
\label{sec:method}
\subsection{Rethink Janus Problem}

Although numerous studies have identified generation bias in diffusion models as a cause of the Janus Problem~\cite{liu2023zero,armandpour2023re,hong2023debiasing,huang2024dreamcontrol}, a comprehensive proof is still lacking, leaving the issue unresolved. This chapter aims to demonstrate that the long-tailed distribution of viewpoints in the training data of diffusion models is the fundamental factor inducing generation bias, and that the resulting imbalance in viewpoint category guidance for 3D models constitutes the essence of the Janus Problem.

\label{sec:rethin_Janus}
\begin{figure*}[htbp]
    \centering
    \includegraphics[width=1.1\linewidth]{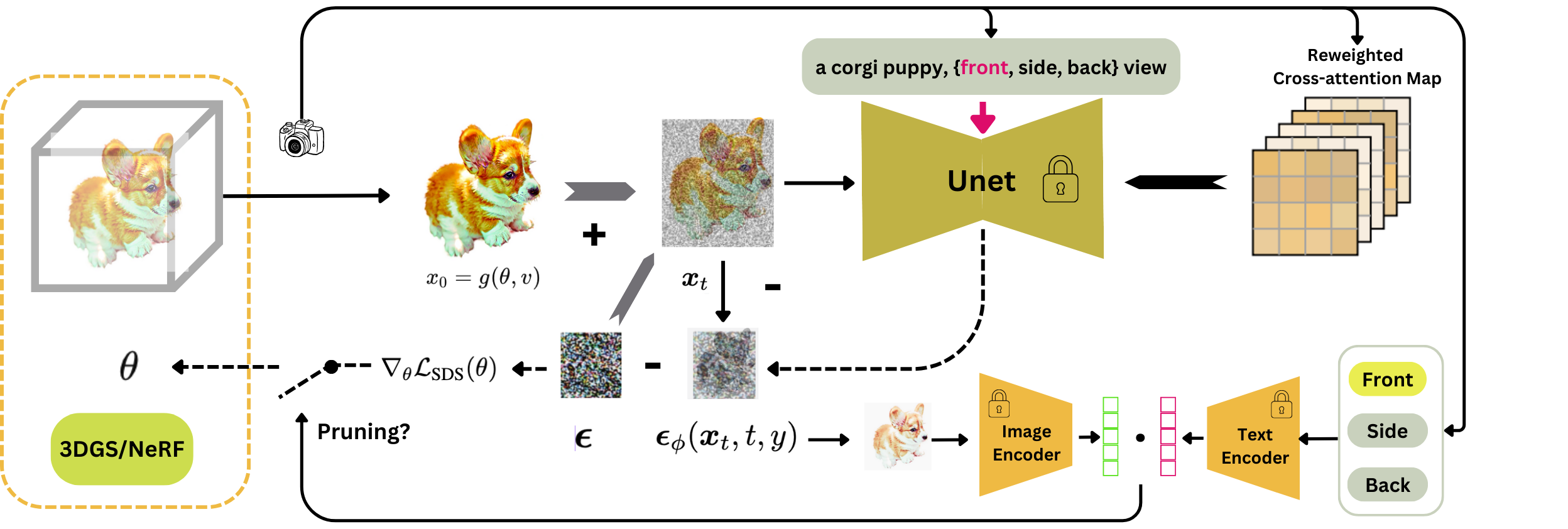}
    \caption{\textbf{Overview of our work.} We adjust cross-attention maps in the diffusion model to enhance focus on viewpoint keywords, guiding the model to generate diverse viewpoints. CLIP guidance prunes incorrect viewpoints during denoising based on a loss threshold, ensuring alignment with specified viewpoints. Decompose the original prompt into object and descriptive words, and optimize the object prompt in the early stages to ensure geometric integrity.}
    \vspace{-2.0em}
    \label{fig:pipeline}
\end{figure*}

\noindent \textbf{Evolution of Data Distribution Over Time.} The diffusion process can be described using both the forward and reverse SDEs. By applying the Fokker-Planck equation to forward SDE~\cite{song2020score}, we can derive:
\begin{equation}
\resizebox{0.47\textwidth}{!}{
$
\frac{\partial p_t(x)}{\partial t} = - \nabla_x \cdot \left( \left( f(x,t) - \frac{1}{2} g^2(t) \nabla_x \log p_t(x) \right) p_t(x) \right), \label{equation5}
$
}
\end{equation}

\noindent where \( p_t(x) \) is the time-evolved data distribution. This equation represents how the original data distribution \( p_0(x) \) changes over time \( t \) from \( 0 \) to \( T \), eventually converging to a standard Gaussian distribution. For the reverse SDE, by letting \( s = T - t \), we obtain a forward process from \( s = 0 \) to \( T \). Assuming the model's predicted score is accurate, the Fokker-Planck equation for the reverse SDE becomes:

\begin{equation}
\resizebox{0.48\textwidth}{!}{
$
\frac{\partial q_s(x)}{\partial s} = - \nabla_x \left[ \left( \frac{1}{2} g^2(t) \cdot \nabla_x \log p_t(x) - f(x,t) \right) \cdot q_s(x) \right], \label{equation6}
$
}
\end{equation}

\noindent which describes the evolution of the Gaussian distribution over time \( s \) from \( 0 \) to \( T \), corresponding to \( t \) decreasing from \( T \) to \( 0 \). This is essentially the reverse of Eq.~\ref{equation5}. 

\noindent \textbf{Revisiting SDS.} The SDS formula proposed by DreamFusion~\cite{poole2022dreamfusion} can be expressed as:
\begin{equation}
\resizebox{0.42\textwidth}{!}{
$
\nabla_{\theta} \mathcal{L}_{\text{SDS}}(\theta) \approx \mathbb{E}_{t,\epsilon,c} \left[ \omega(t) \left( \epsilon_{\phi}(x_t, t, y) - \epsilon \right) \frac{\partial g(\theta, c)}{\partial \theta} \right],  \label{equation7}
$
}
\end{equation}
where \(\epsilon\) represents the 2D rendered image \( x_0 = g(\theta, v) \) of a 3D object, with standard Gaussian noise added at time \( t \). The symbol \( \epsilon_{\phi} \) represents the noise predicted by the U-Net~\cite{ronneberger2015u} for the given prompt and noisy image. Let \( \gamma(t) = \frac{\sqrt{1 - \bar{\alpha}_t}}{\sqrt{\bar{\alpha}_t}} \). From DDPM~\cite{ho2020denoising, song2020denoising}, we have:

\begin{equation}
\bar{x}_0^t = \frac{x_t - \sqrt{1 - \bar{\alpha}_t} \, \epsilon_{\phi}(x_t, t, y)}{\sqrt{\bar{\alpha}_t}},\label{equation8}
\end{equation}

\noindent which implies that the pseudo-GT image can be expressed using the noisy image and the predicted noise. Eq.~\ref{equation7} can be transformed into:
\begin{equation}
\nabla_{\theta} \mathcal{L}_{\text{SDS}}(\theta) = \mathbb{E}_{t,\epsilon,c} \left[ \frac{\omega(t)}{\gamma(t)} \left( x_0 - \bar{x}_0^t \right) \frac{\partial g(\theta, c)}{\partial \theta} \right],  \label{equation9}
\end{equation}

In this context, we interpret SDS as a mechanism that uses images generated from diffusion models~\cite{liang2024luciddreamer}  to guide images rendered by 3D model. The long-tailed distribution of viewpoints in the training data where the majority of images are front-facing introduces bias into the generated images~\cite{sehwag2022generating}, as shown in Figure~\ref{fig:long_tail}. When the generated images exhibit a similarly long-tailed distribution~\cite{hong2023pointcam,han2025enhancing}, the guidance becomes imbalanced, causing the 3D object to overfit to images from the highest-density region.

Moreover, since the Janus problem arises due to incorrect guidance from 2D diffusion image generation, current methods~\cite{poole2022dreamfusion, armandpour2023re, liang2024luciddreamer, liu2023threestudio, wang2024prolificdreamer} try to address this by modifying the original input text based on the rendering angle \( v \) of the current 3D representation. This process involves generating phrases such as prompt + “back view” to optimize the 3D representation by providing directional textual guidance. However, as illustrated in Figure~\ref{fig:attention_map}, we observe that the model’s focus on directional terms diminishes significantly as prompt complexity increases, reducing the effectiveness of this approach. Motivated by these observations, we propose ACG to address these limitations and achieve improved results.

\subsection{Cross-attention Control}
The Janus problem arises from viewpoint bias in generative models. This bias leads to the production of pseudo-GTs \(\bar{x}_0 \) that mis-guide the rendered image \( x_0 = g(\theta, v) \), where \( g \) is the rendering function dependent on parameters \( \theta \) and conditions \( v \).

Cross-attention Control aims to direct the model to generate images from low-probability density regions without altering the training data or fine-tuning the model.

\begin{figure*}[htbp]
    \centering
    \includegraphics[width=1.05\linewidth]{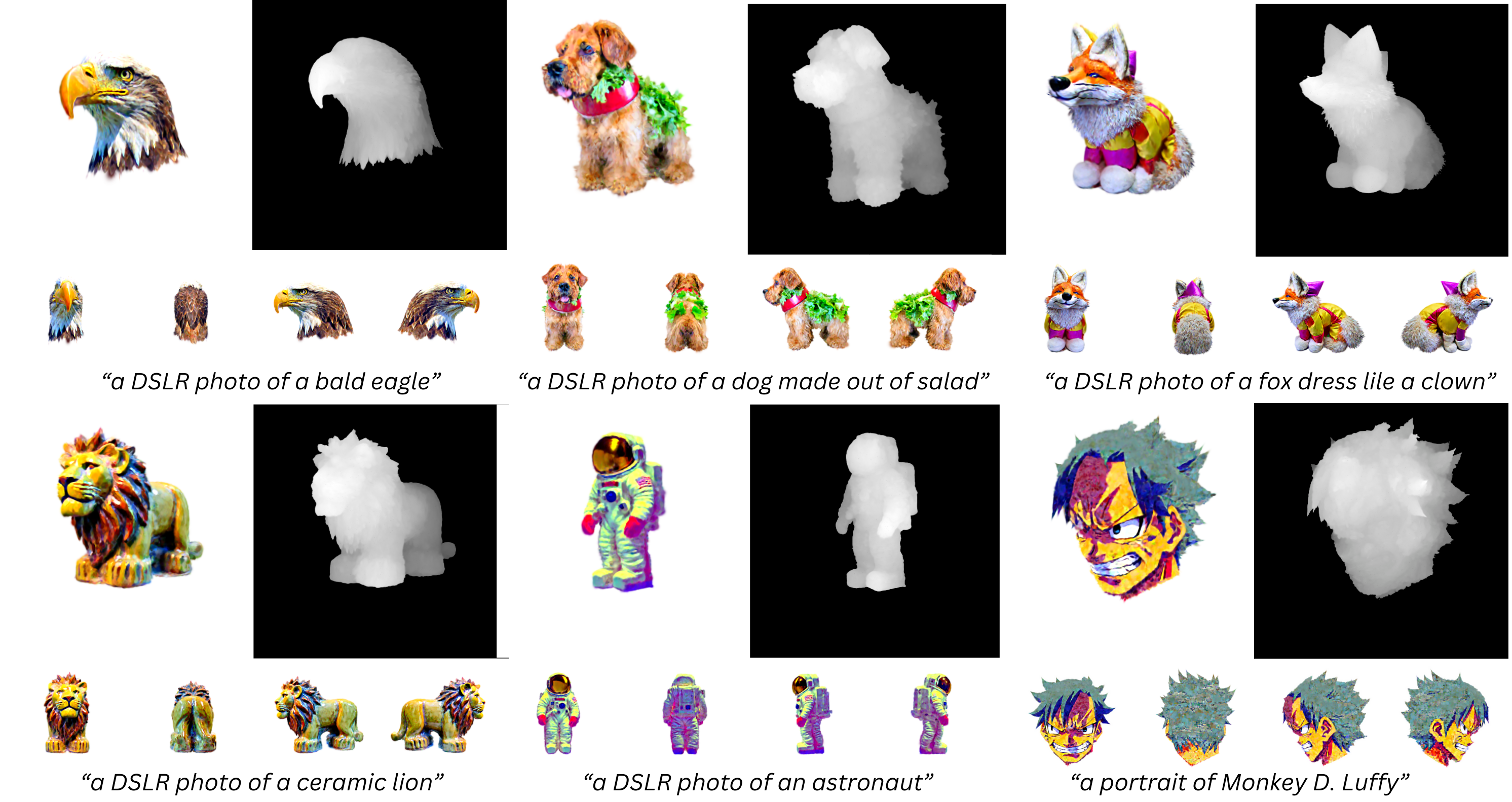}
    \caption{\textbf{Examples generated by ACG.} ACG can enhance the consistency of geometric shapes and the variety of high-fidelity textures in text-to-3D content generation.}
    \vspace{-1.0em}
    \label{fig:main}
\end{figure*}

\noindent \textbf{Cross-Attention in U-Net.} Stable Diffusion~\cite{rombach2022high} serves as our base model, incorporating cross-attention layers in the U-Net's encoding and decoding stages to merge text and image features. The latent image \( z_t \) is transformed via \( \phi(z_t) \) into a query matrix \( Q = l_Q(\phi(z_t)) \), while the text prompt \( P \) is embedded as \( \psi(P) \) and mapped to key and value matrices \( K = l_K(\psi(P)) \) and \( V = l_V(\psi(P)) \), respectively~\cite{hertz2022prompt}. The cross-attention map is calculated as:

\begin{equation}
M = \text{Softmax} \left( \frac{QK^\top}{\sqrt{d}} \right), \label{equation10}
\end{equation}

\noindent where \( d \) is the key vector dimensionality. This map \( M \) weights the value matrix \( V \), producing \( MV \) to update \( \phi(z_t) \).

This mechanism enables the model to focus on spatial regions corresponding to the text~\cite{hertz2022prompt}. We leverage this property to enhance attention on viewpoint-related information in the prompt by controlling \( M \), improving image generation in low-density areas.

\noindent \textbf{Cross-Attention Control.} To alleviate the generation bias of the diffusion model, our method adjusts the cross-attention map \( M \) in the diffusion model to enhance the probability of generating images with correct viewpoints. The map \( M \) aligns latent spatial features with text prompts by comparing the query vector \( Q \) and the key vector \( K \), a mechanism often used in image editing to control generation~\cite{hertz2022prompt, brooks2023instructpix2pix}.

During each noise prediction step in the U-Net of the diffusion process, we identify viewpoint-related keywords in the text prompt: terms that specify spatial viewpoint, such as ``front", ``side", or ``back". These keywords are associated with spatial information that influences the viewpoint of the generated image. We dynamically adjust the attention weights of these viewpoint keywords based on the desired rendering perspective. This adjustment is implemented using the following formula:
\begin{equation} M' = M \odot (1 + \lambda \cdot \mathbf{I}_{\text{keywords}}), \label{equation11} \end{equation}

where \( M \in \mathbb{R}^{N \times T} \) is the original cross-attention map, where \( N \) is the number of spatial locations (latent variable dimensions) and \( T \) is the number of text tokens. The symbol \( \odot \) denotes element-wise multiplication. The coefficient \( \lambda \) adjusts attention weights: positive values enhance, and negative values reduce attention to viewpoint keywords. The indicator matrix \( I_{\text{keywords}} \in \mathbb{R}^{N \times T} \) marks positions corresponding to viewpoint keywords with 1 and others with 0.

\begin{figure*}[t]
    \centering
    \includegraphics[width=0.9\linewidth]{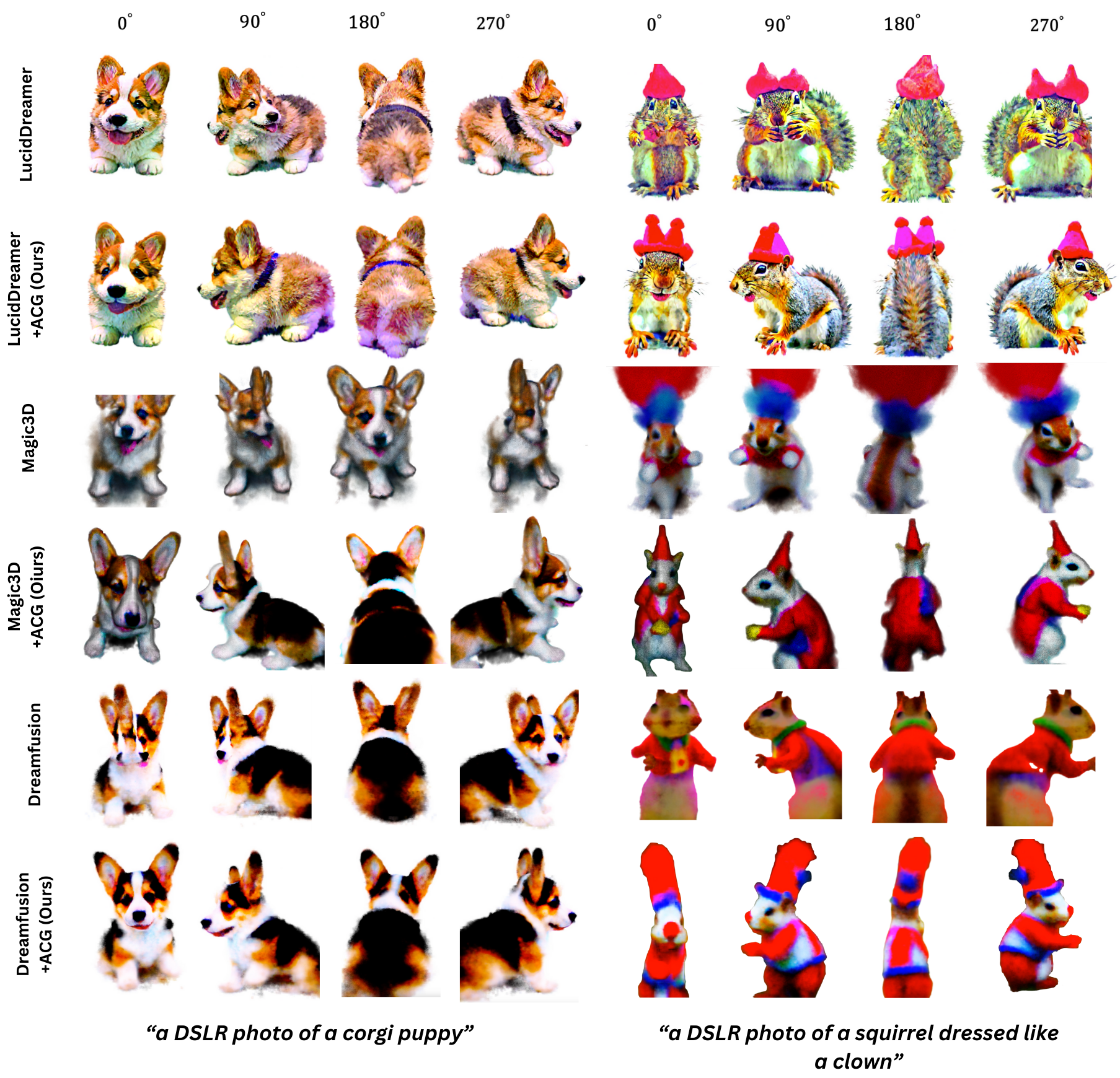}
    \vspace{-0.8em}
    \caption{\textbf{Qualitative comparison with baseline methods in text-to-3D generation.} From top to bottom: Stable-DreamFusion, Magic3D, LucidDreamer, and our ACG-enhanced versions. For each prompt, we show the 0°, 90°, 180,° and 270° views. ACG consistently recovers the missing back geometry while preserving fine texture, and does so without extra optimisation time or VRAM overhead.
}
    \vspace{-1.0em}
    \label{fig:example}
\end{figure*}

By selectively amplifying or suppressing the model's attention to viewpoint-related keywords, we guide the generation toward low-probability density images. This manipulation of the cross-attention map enhances the incorporation of viewpoint information during the generation process, increasing the probability of producing images from the correct viewpoint and effectively addressing the first aspect of the Janus problem.

\subsection{CLIP Pruning}
To mitigate the imbalanced distribution caused by generation bias in diffusion models, we propose CLIP Pruning—a strategy that utilizes CLIP to measure semantic similarity between images and text.


After computing the pseudo-GT image \( \bar{x}_0 \) (Eq.~\ref{equation8}), we encode it into the CLIP image embedding space. Simultaneously, we encode the text prompt \( P_v \), corresponding to the azimuth angle \( v \) of the current view, into the CLIP text embedding space. We compute the CLIP similarity \( \sigma \) between the embeddings to assess semantic alignment:


\begin{equation}
\sigma = \frac{e_{\text{image}} \cdot e_{\text{text}}}{\| e_{\text{image}} \| \| e_{\text{text}} \|}, \label{equation12}
\end{equation}

where \( e_{\text{image}} \) and \( e_{\text{text}} \) represent the embeddings of \( \bar{x}_0 \) and \( P_v \), respectively. The dot product measures semantic alignment, and \( \sigma \) is normalized to a range between \(-1\) and \(1\). If \( \sigma \) falls below a specified threshold \( \tau \), it indicates that the current generation provides incorrect guidance for the 3D model under the specified viewpoint. In such cases, the update step is skipped to prevent the inclusion of misleading guidance that could cause the Janus problem.

\noindent \textbf{Adaptive Threshold.} To dynamically determine the pruning threshold \( \tau \) based on the model's semantic similarity statistics, we introduce an adaptive thresholding mechanism. During the first epoch, we collect CLIP similarity scores \( \sigma \) without pruning to obtain a representative distribution of semantic alignment values. We define \( \tau \) as:

\begin{equation}
\tau = \alpha \cdot \sigma_{\text{min}} + (1 - \alpha) \cdot \sigma_{\text{mean}},
\label{equation13}
\end{equation}

\noindent where \( \sigma_{\text{min}} \) and \( \sigma_{\text{mean}} \) are the minimum and mean values of \( \sigma \) observed during the first epoch. The hyperparameter \( \alpha \in [0, 1] \) adjusts the sensitivity of the pruning mechanism, balancing between a stricter (minimum-based) and a more lenient (mean-based) criterion.

CLIP Pruning mitigates generation bias by eliminating guidance with insufficient semantic view alignment, thereby balancing the distribution of pseudo-GT viewpoints during training. This pruning strategy rebalances the generation bias to some extent, leading to a uniform sampling distribution of viewpoints in 3D generation.

\subsection{Coarse-to-Fine Optimization}
\noindent
We observed that increasing prompt complexity reduces the diffusion model's focus on individual words, particularly those conveying viewpoint information, as shown in Fig.~\ref{fig:attention_map}. To address this, we propose a two-stage optimization approach that enhances the model's attention to essential elements.

We first divide the input prompt into an object prompt (main subject) and a description prompt (additional characteristics). In the initial stage, we focus on the object prompt by adjusting cross-attention maps and applying CLIP pruning to amplify attention to spatial viewpoint terms. This ensures the generated 3D model accurately captures the core geometry and specified viewpoint.

Then, we introduce the description prompt to refine the appearance and add details, enhancing realism. By progressively refining from basic geometry to intricate details, we mitigated the loss of viewpoint information that occurs as prompts become increasingly complex.

\section{Experiments}
\label{sec:exp}
\subsection{Text-to-3D Generation}
To demonstrate the effectiveness of our approach, we conduct two primary experiments. First, we evaluate the reduction of the Janus Problem in text-to-3D generation through both qualitative and quantitative comparisons, showing that our method significantly mitigates this issue. Second, we conduct ablation studies to assess the contribution of each module within our approach.

For the cross-attention map parameter  $\lambda$, its value is set based on the length of the prompt, scaled by a factor of 10. In staged optimization, we typically divide the prompt into two segments, each corresponding to approximately half of the optimization process.

\noindent \textbf{Qualitative Comparison.}
We compare our model with the current state-of-the-art baseline implemented by stable-Dreamfusion~\cite{stable-dreamfusion}, Three-Studio~\cite{liu2023threestudio} and LucidDreamer~\cite{liang2024luciddreamer}. All methods utilize Stable Diffusion v1.4 for distillation, and experiments are conducted on an NVIDIA 4090 GPU for fair comparison. As shown in Figure \ref{fig:example}, our method demonstrates noticeable improvements in handling the multi-head problem.

\noindent \textbf{Quantitative Comparison.}
Previous research in 3D generation lacks standardized metrics. Existing works typically use 2D image evaluation metrics to assess the quality or prompt relevance of generated content. However, these metrics are insufficient for evaluating the Janus problem. To assess the consistency of 3D geometry, we generate 3D objects using 20 randomly selected text prompts from the DreamFusion prompt dataset. For each 3D object,  we count the instances of inconsistent 3D content produced by each approach, defining this count as the Janus Problem Rate (JR). As shown in Table~\ref{table1}, our method significantly reduces the multi-head problem.
\begin{table}[t]
\centering
\begin{tabular}{lcc}
\toprule
\textbf{Method} & \textbf{JR(\%)$\downarrow$} 
\\
\midrule
LucidDreamer~\cite{liang2024luciddreamer} & 80.00 \\
LucidDreamer+ACG (Ours) & \textbf{30.00}\\
\midrule

Magic3D~\cite{lin2023magic3d} & 75.00  \\
Magic3D+ACG (Ours) & \textbf{35.00}  \\
\midrule

DreamFusion~\cite{poole2022dreamfusion} & 80.00 \\
DreamFusion+ACG (Ours) & \textbf{35.00} \\

\bottomrule
\end{tabular}
\caption{\textbf{Comparison of geometric consistency between baseline methods and ACG-enhanced versions.} Lower JR indicates fewer Janus outcomes. ACG reduces JR across all baselines.}
\vspace{-0.8em}
\label{table1}
\end{table}

\subsection{Ablation Study}
\begin{figure}[t]
    \centering
    \includegraphics[width=1\linewidth]{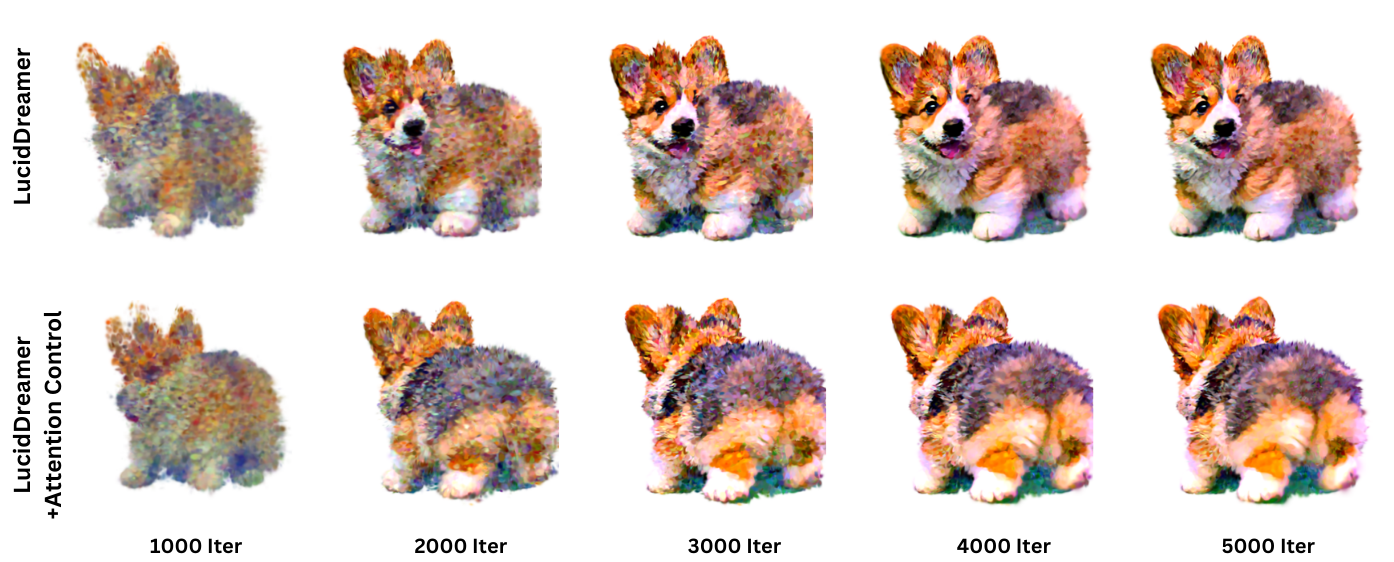}
    \vspace{-1.2em}
    \caption{\textbf{Ablation study on controlling cross attention map.} Controlling Cross Attention can suppress the generation of front views when the rendering viewpoint is at the back view.}
    \label{fig:Ablation1}
    \vspace{-0.4em}
\end{figure}

\begin{figure}[t]
    \centering
    \includegraphics[width=1\linewidth]{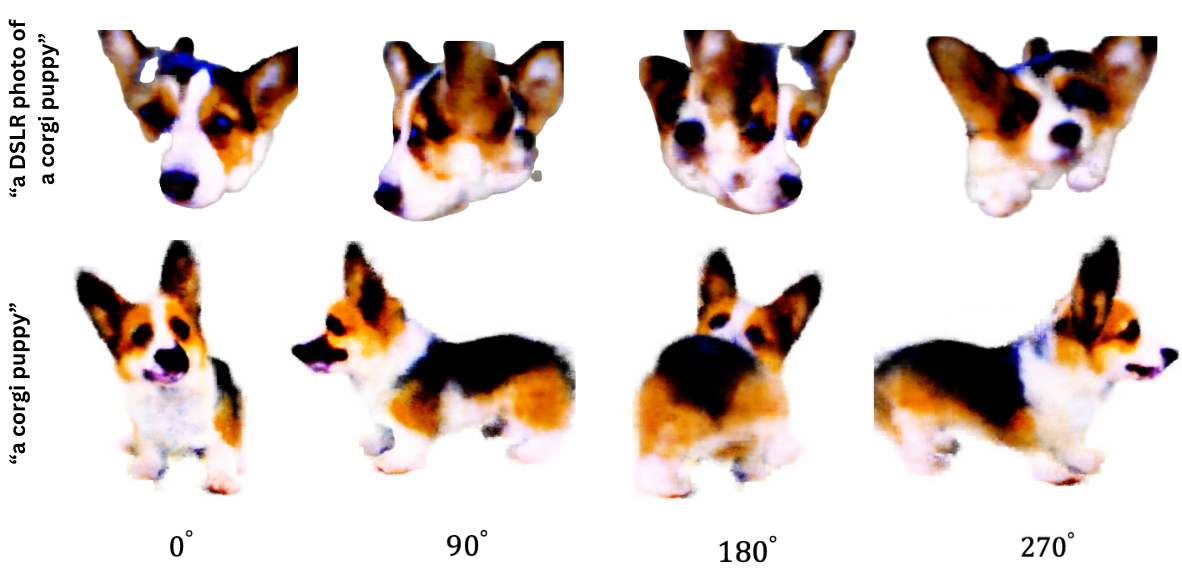}
    \vspace{-1.2em}
    \caption{\textbf{Ablation study on coarse to fine optimization based on prompt.} For Dreamfusion~\cite{stable-dreamfusion} with seed=2, the top row uses the original prompt ``a DSLR photo of a corgi puppy'', and the bottom row only uses ``a corgi puppy''.}
    \label{fig:Ablation2}
    \vspace{-1.0em}
\end{figure}

\begin{figure}[t]
    \centering
    \includegraphics[width=1\linewidth]{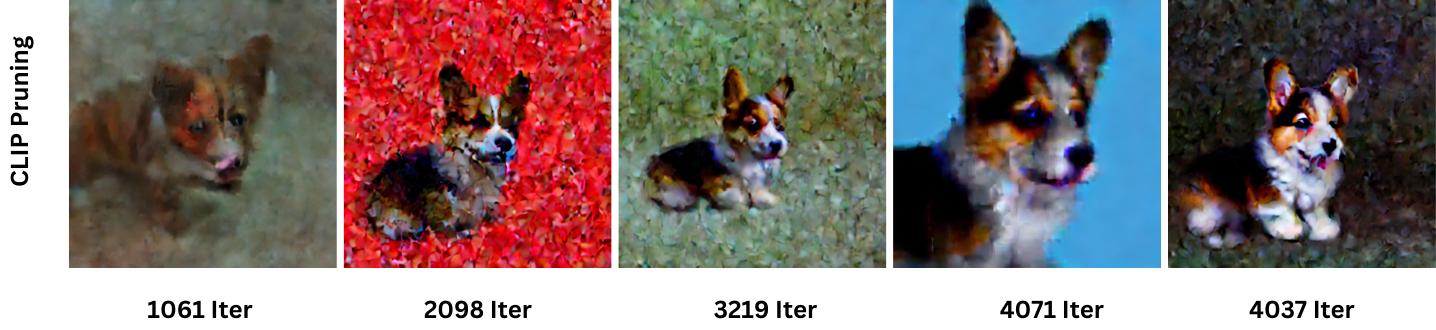}
    \vspace{-0.8em}
    \caption{\textbf{Ablation study on CLIP Pruning.} These pseudo-Ground-Truth images identified by CLIP as front-facing are pruned in the back viewpoint range of [120, 240] degrees. This ensures only back-view images are retained, enhancing viewpoint consistency in the generated 3D representations.}
    \label{fig:Ablation3}
    \vspace{-1.0em}
\end{figure}

To verify the effectiveness of each component in our method, we take the generation of ``a DSLR photo of a corgi puppy'' as an example and compare the results with and without our design in Figures~\ref{fig:Ablation1}, \ref{fig:Ablation2}, and \ref{fig:Ablation3}.

\noindent \textbf{Controlling Cross-attention Map.}
In Figure \ref{fig:Ablation1}, we compare intermediate renderings of the baseline and cross-attention-only 3D representations, with the rendering viewpoint fixed at a horizontal angle of 135 degrees. It can be observed that both methods display prominent facial features in the early stages. However, by controlling the cross-attention map to increase the UNet’s focus on the term “back” from this angle, facial features are suppressed in the back view, resulting in a 3D representation that clearly exhibits back characteristics.

\noindent \textbf{Coarse-to-Fine Optimization.} In Figure \ref{fig:Ablation2}, we compare 3D-rendered images generated from the original prompt, “a DSLR photo of a corgi puppy”, with those from a simplified prompt, “a corgi puppy”, using the same seed. DreamFusion may produce unrealistic 3D shapes when using the original prompt, whereas the simplified prompt can somewhat reduce optimization difficulty, leading to a better initialization in the early stages. This initial simplification establishes a foundation for introducing more complex descriptive terms later in the process. However, despite this improvement, the optimized result from the simplified prompt still displays unintended facial features on the back.

\begin{figure}[t]
    \centering
    \includegraphics[width=0.8\linewidth]{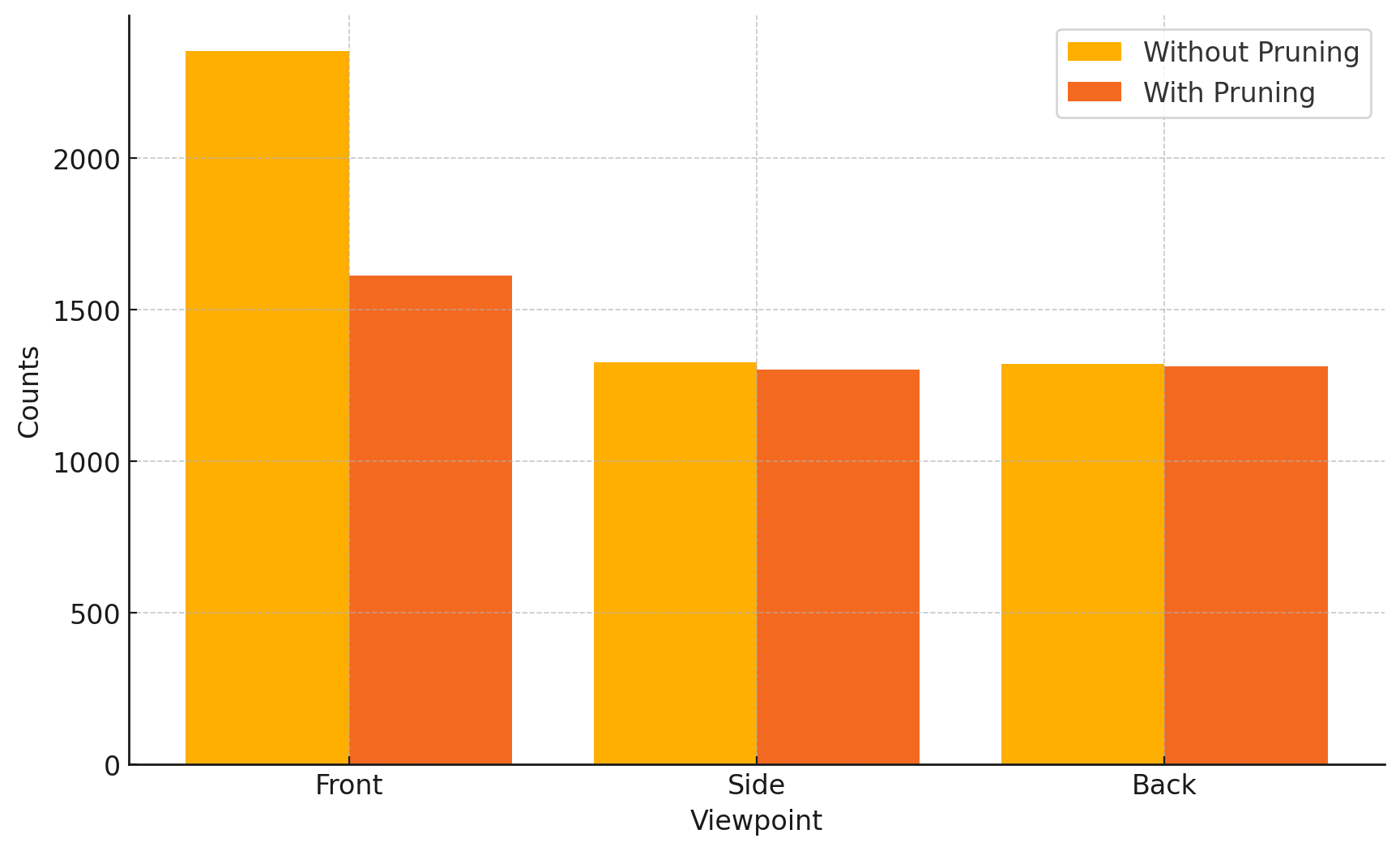}
    \vspace{-0.8em}
    \caption{\textbf{Distribution of viewpoints without and with pruning.} The chart compares counts of pseudo-GT viewpoints during training under two conditions: without pruning (approximated ratio 2:1:1) and with pruning (approximated ratio 1.2:1:1).}
    \vspace{-1.0em}
    \label{fig:Ablation5}
\end{figure}

\noindent \textbf{CLIP Pruning.}
In Figure \ref{fig:Ablation3}, we demonstrate the effectiveness of CLIP in pruning incorrect optimizations. We define the back range of the object as [120, 240] degrees. When the rendering viewpoint falls within this range, if the pseudo-GT image from the Diffusion Model is identified as front-facing, that optimization is pruned. Figure \ref{fig:Ablation3} shows the pseudo-GTs images pruned at different iterations, all displaying prominent facial features. 

\noindent \textbf{Viewpoint Rebalancing.} Additionally, we analyzed the impact of CLIP Pruning on the balance of pseudo-GT viewpoint distribution during training. As shown in Figure~\ref{fig:Ablation5}, we expected an even distribution across the three viewpoints (front, side, and back), as we randomly sample the three viewpoints with equal probability. However, without the pruning, the pseudo-GT viewpoint reveals a substantial front-view bias with a ratio of approximately 2:1:1. After applying our pruning technique, we observed a significant improvement in balance, achieving a more equitable distribution of 1.2:1:1. This adjustment aligns the pseudo-GT viewpoint more closely with our expectation and is critical for enhancing model's capacity to overcome Janus problem.

\section{Conclusion}

In this paper, we analyze the Janus Problem in text-to-3D generation and identify it as viewpoint generation bias. This bias stems from the long-tailed distribution of training data in diffusion models and the increased complexity of input prompts, leading to incorrect guidance in 3D optimization. 
To address this issue, we introduce our method, ACG, to alleviate the viewpoint generation bias. Specifically, ACG controls the cross-attention map and utilizes CLIP to guide and supervise the diffusion model's generation process. Our approach effectively mitigates the Janus Problem without requiring fine-tuning. Notably, our method can be integrated as a plugin into other text-to-3D methods to significantly reduce the occurrence of the Janus Problem without affecting their generation speed.

\noindent \textbf{Limitation.} While our method increases the likelihood of generating viewpoints with low probability density in diffusion models, it still struggles with certain extremely rare object categories. Additionally, CLIP, being a general-purpose cross-modality model, is not specifically designed for viewpoint recognition and thus provides only coarse guidance for viewpoints.

\noindent \textbf{Future Work.} Having identified the Janus Problem as a generation bias caused by long-tailed distributions, future work could pursue two complementary directions: (i) fine-tuning the CLIP on the noise-corrupted renderings that arise during diffusion, so its similarity signals remain balanced across both frequent and rare viewpoints, and (ii) reshaping the SDS guidance by injecting tail-view exemplars, explicitly drawing the optimisation gradients toward the low-density regions of viewpoint space.

{
 \small
 \bibliographystyle{ieeenat_fullname}
 \bibliography{main}
}


\end{document}